\documentclass{article}
\usepackage{ijcai24}

\usepackage{amsmath} % assumes amsmath package installed
\DeclareMathOperator*{\argmax}{arg\,max}

\usepackage{amssymb}  % assumes amsmath package installed
\usepackage{courier}  % DO NOT CHANGE THIS
\usepackage{graphicx} % DO NOT CHANGE THIS
\usepackage{soul}
\usepackage{url}
\usepackage{bbm}
\usepackage{balance}
\usepackage[utf8]{inputenc}
\usepackage{mathtools}
\usepackage{booktabs}
\usepackage{algorithm}
\usepackage{algorithmic}
\usepackage{graphicx}
\usepackage{hyperref}
\usepackage{multirow}
\usepackage{rotating}
\usepackage{wasysym}
\usepackage{caption}
\usepackage{subcaption}
\usepackage[dvipsnames,table,xcdraw]{xcolor}

\usepackage{enumitem}
\usepackage{notes2bib}
\usepackage{pdfpages}

\newcommand\blfootnote[1]{%
  \begingroup
  \renewcommand\thefootnote{}\footnote{#1}%
  \addtocounter{footnote}{-1}%
  \endgroup
}

\usepackage[switch]{lineno}

\linenumbers

\title{\LARGE \bf
Toward Informed AV Decision-Making: Computational Model of Well-being and Trust in Mobility}

\author{
Zahra Zahedi
\and
Shashank Mehrotra\and
Teruhisa Misu\And
Kumar Akash\\
\affiliations
Honda Research Institute USA. Inc\\
\emails
\{zahra\_zahedi, shashank\_mehrotra, tmisu, kakash\}@honda-ri.com
}
\begin{document}
\nolinenumbers
\maketitle

\begin{abstract}
% For future human-autonomous vehicle (AV) interactions to be effective and smooth, human-aware systems that analyze and harmonize human's needs with automation decisions are required. To achieve this, it is critical to design systems that account for human cognitive states. We present a novel computational model in the form of a Dynamic Bayesian Network (DBN) that can infer the cognitive state of both AV users and other road users and incorporate this information into its decision-making process. Specifically, our model incorporates ``well-being'' of both an AV user and an interacting road user as cognitive states along with trust. Our DBN models infer beliefs over the AV user’s event-to-event latent wellbeing, trust and intention states and possibly other road user's wellbeing based on the observed interaction experiences. Using datasets collected from an interaction study, we refine the parameters and empirically assess the model performance. We then extend our model as a causal inference model (CIM) for informed AV decision-making, allowing the AV to enhance user's well-being and trust while balancing them with ego's cost and well-being of the interacting road user. These evaluation highlight our model in prediction accuracy and informed decision making.

For future human-autonomous vehicle (AV) interactions to be effective and smooth, human-aware systems that analyze and align human needs with automation decisions are essential. Achieving this requires systems that account for human cognitive states. We present a novel computational model in the form of a Dynamic Bayesian Network (DBN) that infers the cognitive states of both AV users and other road users, integrating this information into the AV's decision-making process. Specifically, our model captures the ``well-being'' of both an AV user and an interacting road user as cognitive states alongside trust. Our DBN models infer beliefs over the AV user’s evolving well-being, trust, and intention states, as well as the possible well-being of other road users, based on observed interaction experiences. Using data collected from an interaction study\footnote{The study was approved by the Bioethics Committee in Honda R\&D (approval code: 99HM-065H)}, we refine the model parameters and empirically assess its performance. Finally, we extend our model into a causal inference model (CIM) framework for AV decision-making, enabling the AV to enhance user well-being and trust while balancing these factors with its own operational costs and the well-being of interacting road users. Our evaluation demonstrates the model’s effectiveness in accurately predicting user's states and guiding informed, human-centered AV decisions. 
\blfootnote{This is the author’s version of the work. It is posted here for your personal use. Not for redistribution. The definitive Version of Record is published in Proceedings of the 34th International Joint Conference on Artificial Intelligence (IJCAI-25), Human-Centred Artificial Intelligence Track.}\looseness=-1

% We implement and evaluate the model based on the interaction between a self-driving scooter as the AV and delivery robots as other road users. We first conduct a user study to gain insights into the correlation of various variables, which we use to inform the development of our DBN-based model of trust and well-being. Then, we train and test our model, which can infer beliefs over the cognitive states of humans and other users using the collected data. Finally, we extend our model as a causal inference model (CIM) for informed AV decision-making, allowing the AV to enhance user's well-being and trust while balancing them with ego's cost and well-being of the interacting road user.
\end{abstract}
% \vspace{-9pt}
\section{INTRODUCTION}
With the proliferation of autonomous vehicles (AVs), including cars and even smaller autonomous vehicles such as delivery robots and drones, humans are bound to encounter AVs in more places and in different forms. However, existing research on automation techniques  overlooks the environmental and social implications while developing these systems \cite{zhuge2020toward}. For example, recently, some of these new mobility modes that had been considered convenient and climate-friendly have received a public perception of being dangerous and a nuisance. For example, Paris referendum, 89\% of voters supported a ban on electric scooters \cite{nouvian2023paris}. One of the primary reasons for such public sentiment is that these shared mobility modes only consider the basic mobility needs of the users with disregard for their well-being, satisfaction, and positive relationship with other road users \cite{ettema2011satisfaction}. Given the recent convergence between automated vehicle (AV) technology and shared mobility, new small-scale shared automated vehicle tests are beginning to develop around the world \cite{stocker2017shared}. Therefore, it becomes even more critical for these AVs to holistically consider the needs of the users as well as nearby road users.
% rust in the AV plays a critical role in shaping their response to the vehicle's actions. T
A potential paradigm to achieve this is to account for human cognitive states while making AV decisions. Studies have established a strong association between driving behavior and well-being \cite{harris2014prosocial}. Similarly, a user’s trust influences how users perceive the AV’s intentions and reliability, which in turn affects their willingness to accept its decisions \cite{lee2004trust}. Thus, it is necessary for the AV to quantitatively measure users' well-being and trust to better anticipate the impact of its actions on both its user and others on the road. We propose to focus on \emph{well-being} as a cognitive state to account for holistic user needs that include situational satisfaction as well as positive relation with other road users, and to ensure successful interaction between the user and their AV, we incorporate user's \emph{trust} in the AV.

In this paper, we propose a Bayesian-based model that allows for the inference of well-being and trust without disrupting the user's behavior. The proposed model can continuously maintain an estimate of the user's cognitive states and update it based on the user's and other road users' latent states. We use Lee and See's definition of user's trust as their attitude toward AV/others that they will help them achieve their goal of driving safely in a situation characterized by uncertainty and vulnerability \cite{lee2004trust}. Well-being is a multidimensional construct and constitutes several factors \cite{seligman2002positive}. For the mobility context, we define well-being to comprise of positive social interactions, satisfaction with travel, trust in other road users, and general well-being of the user \cite{radzyk2014validation}. Furthermore, our work introduces the novel application of incorporating inferred well-being into decision-making processes, a unique contribution that has not been explored in previous research.
The main contributions of the paper include: \looseness = -1
\begin{enumerate} 
\item Development of a quantitative computational model of trust and well-being based on a dynamic Bayesian network (DBN) that can infer user's trust, well-being, and intention. The model was then trained and evaluated using the collected user study data. 
% \item We first conduct an extensive observational user study to gather empirical data and insights to inform the development of our model. 
% \item Using the collected data and domain knowledge, we develop a quantitative computational model represented as Bayesian Dynamic network (DBN) of trust and well-being. The model formulates Bayesian beliefs over the states of trust, well-being, and intention of the user, as well as the well-being of other road users. 
\item Inference of optimal decision-making policies through interaction to achieve desired objectives, such as maximizing the user's well-being and trust as well as optimizing a trade-off between the well-being of the user, other road users, and AV costs. 
\end{enumerate}

\section{Related work}
The consideration of cognitive states in human-automation interactions has been an emerging research focus, particularly as autonomous systems become more integrated into everyday environments. Several studies have examined the importance of understanding cognitive states, such as trust, attention, workload, and situational awareness to ensure effective human-automation collaboration \cite{akash2020toward,azevedo2020context,wu2022toward}. These studies emphasize the need for systems that can interpret and respond to these states, with the aim of enhancing user experience and safety. \looseness = -1

One key challenge is the dynamic nature of cognitive states, which evolve over time based on the interaction context. 
Bayesian inference and modeling have been widely employed in various domains for modeling cognitive states, including workload \cite{koppol2021interaction,luo2019toward,guhe2005non}, 
trust \cite{xu2015optimo,guo2020modeling,soh2020multi},  
% trust \cite{xu2015optimo,guo2020modeling},  
distraction \cite{liang2007nonintrusive,liang2014hybrid,zahedi2022modeling}, 
% distraction \cite{liang2014hybrid,zahedi2022modeling}, 
and emotion \cite{ong2019computational}.
% , among others \cite{kangasraasio2019parameter}. 
These approaches have yielded valuable insights into human behavior and decision-making processes in diverse domains 
\cite{mahmood2024designing,zahedi2023trust,xu2016maintaining,luo2021workload,deo2019looking}. \looseness=-1
% \cite{zahedi2023trust,xu2016maintaining,luo2021workload}.

In addition to the computational modeling of cognitive states, the relationship between driving and well-being has been an area of research. Various studies have explored how factors like stress, fatigue, and emotional states affect driver behavior and safety, providing indirect evidence of the connection between cognitive states and well-being. For example, levels of well-being are correlated with driving performance \cite{hu2013negative,bowen2019drive}. %Economists have argued that future development should additionally consider individuals' well-being \cite{stiglitz2009report}.  
and levels of driving violations \cite{isler2017life}. Also, prosocial driving behavior promotes cooperation with other road users and reduces incidents of aggressive and stressful driving \cite{harris2014prosocial}. While these studies contribute valuable insights, they typically rely on self-reported measures or physiological indicators to assess well-being. Self-report questionnaires, such as \cite{radzyk2014validation,friman2013psychometric} and physiological measurements, such as \cite{sauer2019empirical,halkola2019towards} are commonly used to assess subjective states of well-being. However, these measures are potentially distracting or intrusive; they may not be practical for real-time decision-making in safety-critical environments. Mehrotra et al. explored the factors impacting well-being and trust and proposed a support vector machine model to understand these factors \cite{mehrotra2023wellbeing}. However, to create a comprehensive model, we require an informed model that incorporates cognitive structures and accounts for the dynamic nature of well-being and trust.\looseness =-1
% \vspace{-5pt}
\section{Problem Formulation}
Our model is predicated on three key relationships: (1) relationship between the user (denoted as $E$)  and their AV (i.e. $R$), (2) relationship between the AV and the other road user (denoted as $O$), and (3) relationship between the user and the other road user. We consider a dyadic bi-directional interaction that involves possible symmetric actions. At each interaction, either the other road user or AV can contribute toward the other by an accommodative action, and the other is the receiver of that. Accommodative action can be choosing a positive prosocial action toward others ($R_+$ for AVs as action contributors, or $O_+$ for others contributing to the action) or not ($R_-/O_-$). When the AV is the contributor, the user of the AV might have an intention for accommodative action toward the other ($I_+$ as positive intention toward accomodative action or not $I_-$) that may or may not align with the one the AV chooses ($Al_1$ or $Al_0$ respectively). Therefore, users well-being, trust, and their action-alignment can affect the relationship between the user and the AV.\\
Formally, the goal of this work is to infer the degree of user's wellbeing $w_k \in [0 \hspace{5pt} 1]$ (where $0$ is the lowest and $1$ is the highest), trust on the AV $t_k \in [0 \hspace{5pt} 1]$ and the intention towards others $i_k \in \{I_{+}, I_{-}\}$, as well as other's wellbeing $w^O_k \in [0 \hspace{5pt} 1]$, at each interaction event $k = 1: K$. We tackle this problem by relating the latent states to observable factors of AV's and others accommodative actions ($a^R_k \in \{R_{+}, R_{-}\}$ and $a^O_k \in \{O_{+}, O_{-}\}$ respectively). Since at any interaction event, the user intention might not necessarily be the same as the AV action; we consider an action-alignment state, represented by $al_k \in \{Al_0, Al_1\}$ where $Al_0$ indicates not-aligned and $Al_1$ indicates aligned with the user's intention.
\subsection{Interaction Models}
We formalize the interaction model as two Dynamic Bayesian Networks (DBN). We could have modeled the interaction as a 2-step Temporal Bayesian Network (2TBN), in which the $O$ and $R$ alternatively contribute an accommodative action. However, in order not to limit the framework to alternative interaction between others and AV as an action contributor, we modeled each interaction as a separate DBN, one for when the AV is taking an action (R-DBN, or $R$ contributor DBN), and the other for when the other is taking an action (O-DBN, or $O$ contributor DBN).\\
At each event $k$, our model treats the state of user's latent states and others' states as random variables and maintains belief distributions based on various factors of the interaction experience. Our R-DBN model relates the user’s latent states $x^E_k = \{w_k, \hspace{5pt}t_k,\hspace{5pt} i_k\}$ causally to the AV's action $a^R_k$, other's action $a^O_k$ and user's intention alignment state with the AV $al_k$, and The O-DBN relates the other's latent state $x^O_k = w^O_k$ to AV's action $a^R_k$, their action $a^O_k$. 
Links to each factor in these Bayesian models are quantified as a conditional probability distribution (CPD). How the user's states $x^E_k$ expected to change given the current AV's action $a^R_k$ and action alignment $al_k$ and possible recent others's actions $a^O_{k-1}$ is reflected as follows when AV is action contributor and when other is action contributor respectively:
\begin{eqnarray}
x^E_k \sim Prob(x^E_k|x^E_{k-1}, a^R_k, al_{k})
\end{eqnarray}
and
\begin{eqnarray}
x^E_k \sim Prob(x^E_k|x^E_{k-1}, a^O_{k-1})
\end{eqnarray}
Similarly, for other's state $x^O_k$ the expected change when the AV is contributing to the action is represented as 
\begin{eqnarray}
x^O_k \sim Prob(x^O_k|x^O_{k-1}, a^R_k)
\end{eqnarray}
When others are contributing to an action as
\begin{eqnarray}
x^O_k \sim Prob(x^O_k|x^O_{k-1}, a^O_{k})
\end{eqnarray}
% \begin{eqnarray}
% P(x^E_k,x^E_{k-1}, a_k, al_{k}, a^O_{k-1}) \coloneqq Prob(x^E_k|x^E_{k-1}, a_k, al_{k}, a^O_{k-1})
% \end{eqnarray}
% Similarly, for other's state $x^O_k$ the expected change when the AV is contributing the action is represented as 
% \begin{eqnarray}
% P(x^O_k,x^O_{k-1}, a_k, a^O_{k-1}) \coloneqq Prob(x^O_k|x^O_{k-1}, a_k, a^O_{k-1})
% \end{eqnarray}
% And when other's are contributing an action as
% \begin{eqnarray}
% P(x^O_k,x^O_{k-1}, a_{k-1}, a^O_{k}) \coloneqq Prob(x^O_k|x^O_{k-1}, a_{k-1}, a^O_{k})
% \end{eqnarray}
\begin{figure}[t]
    \centering
    \includegraphics[width=.3\textwidth]{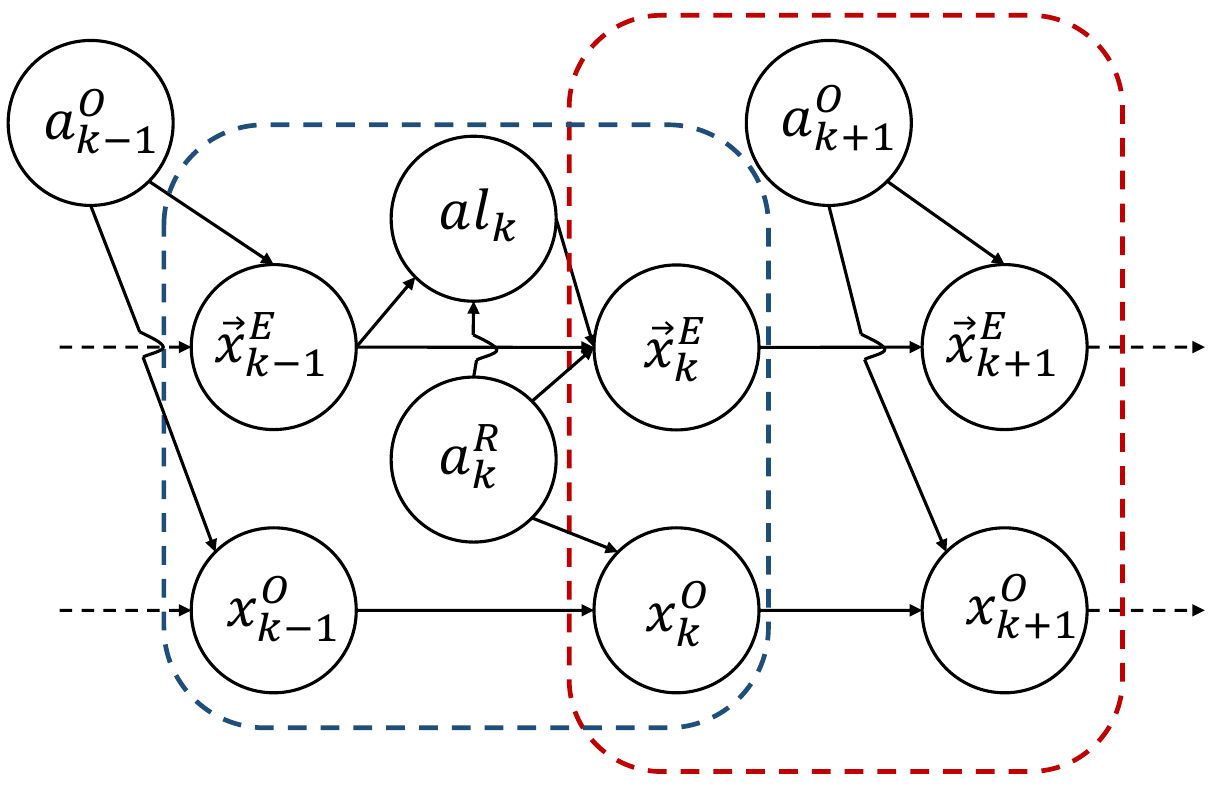}
    \caption{R-DBN (blue) and O-DBN (red) in general form in alternative order.}
    \label{fig:graph} \vspace{-1.8em}
\end{figure}
This probabilistic representation allows us to infer the expected human states and other's state at any given event, as well as the level of uncertainty associated with each estimate. The general graphical model of the casual and evidential variable interaction are shown in Figure \ref{fig:graph}. \looseness = -1

\subsection{Bayesian Inference and Prediction}
The models allow us to estimate the probabilistic belief over the user’s latent states $x^E_k$ at event $k$. Inference is performed by computing the posterior distribution of \( x^E_k \) or $x^O_k$ given past observations. Using Bayesian filtering, we recursively compute
\begin{eqnarray}
    P(x^E_k | \text{evidence}_{1:k}) \propto P(\text{evidence}_k | x^E_k) \nonumber\\
    \times\sum_{x^E_{k-1}} P(x^E_k | x^E_{k-1}, a^R_k, al_k, a^O_{k-1}) P(x^E_{k-1} | \text{evidence}_{1:k-1})
\end{eqnarray}

where \( \text{evidence}_k \) consists of observed variables such as the AV’s action \( a^R_k \), others’ actions \( a^O_k \), and the alignment state \( al_k \).  

For prediction, given an initial belief \( P(x^E_0) \), the future state can be estimated by marginalizing over latent states:

\begin{eqnarray}
    P(x^E_{k+1} | \text{evidence}_{1:k}) = \nonumber\\
    \sum_{x^E_k} P(x^E_{k+1} | x^E_k, a^R_{k+1}, al_{k+1}, a^O_k) P(x^E_k | \text{evidence}_{1:k})
\end{eqnarray}

This enables forward simulation of potential future states, allowing the model to anticipate user behavior under different interaction scenarios.

% We conducted an observational study to collect interaction experiences towards modeling trust relationships. This
% study further yielded pragmatic insights about trust and its
% related constructs, for asymmetric human-robot teams.

% Given the models, we can then perform inference on each variable by querying the propagated belief at event $k$ based on past experiences. Given the structure of our R_DBN and O_DBN models, inference involves computing the posterior distribution over the hidden states given the observed actions and interaction variables. To estimate the latent state distribution at each event  $k$, we employ Bayesian inference, which allows us to update our belief over $x_^E_k$ or $x^O_k$ given new evidence. Specifically, using the Bayesian filtering approach, the posterior distribution over 
% $x_^E_k$ is recursively computed as:
\section{Observational Study and Model Learning}
We adopt a data-driven approach to refine the relationships in our model by conducting an observational study in a controlled setting. This empirical study allow us to analyze the dependencies between latent states and observed actions, providing insights that helped parameterize our DBN models based on real-user data.  \\
Once the structural relationships are established, we parameterize the model by estimating the conditional probability distributions (CPDs) for each variable given its parent nodes. To achieve this, we employ Bayesian parameter estimation, incorporating prior knowledge through a Dirichlet prior with a uniform hyperparameter \( \alpha \). \looseness=-1
\begin{figure*}[t]
    \centering
    \begin{subfigure}[b]{0.15\textwidth}
        \centering
        \includegraphics[width=\textwidth]{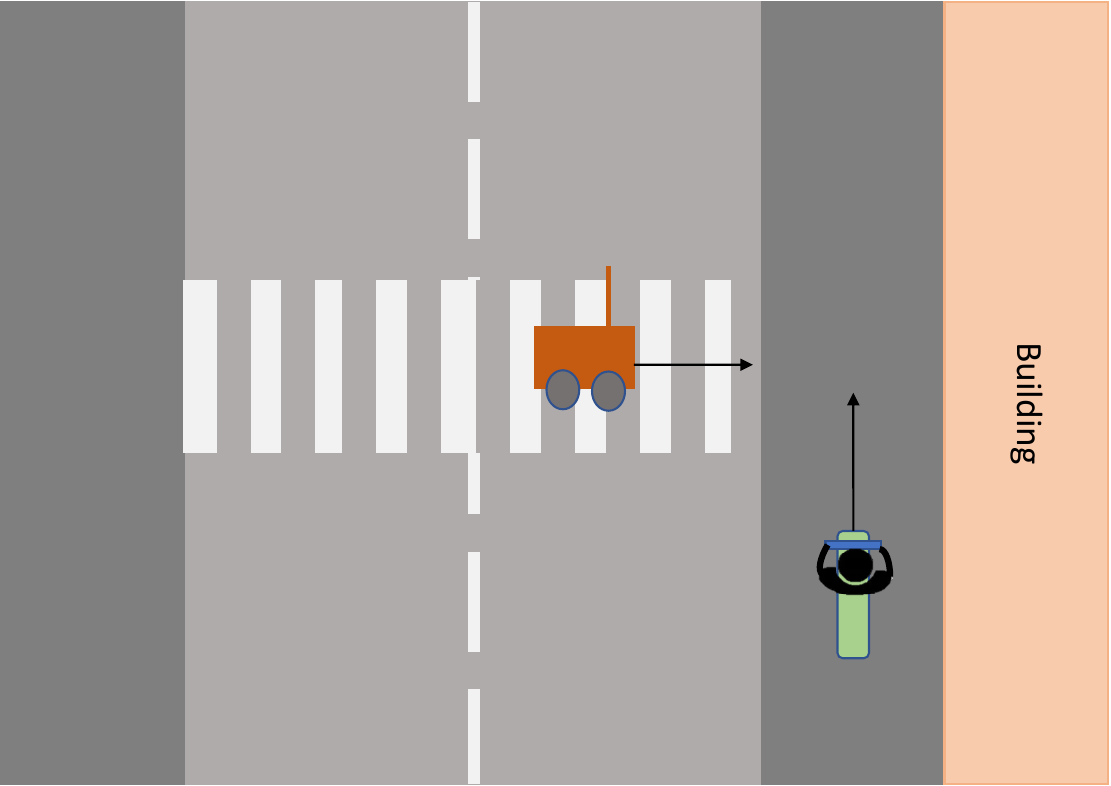}
        \caption{Scenario $S1$}\label{fig:s1}
    \end{subfigure}
    \hspace{20pt}
    \begin{subfigure}[b]{0.15\textwidth}
        \centering
        \includegraphics[width=\textwidth]{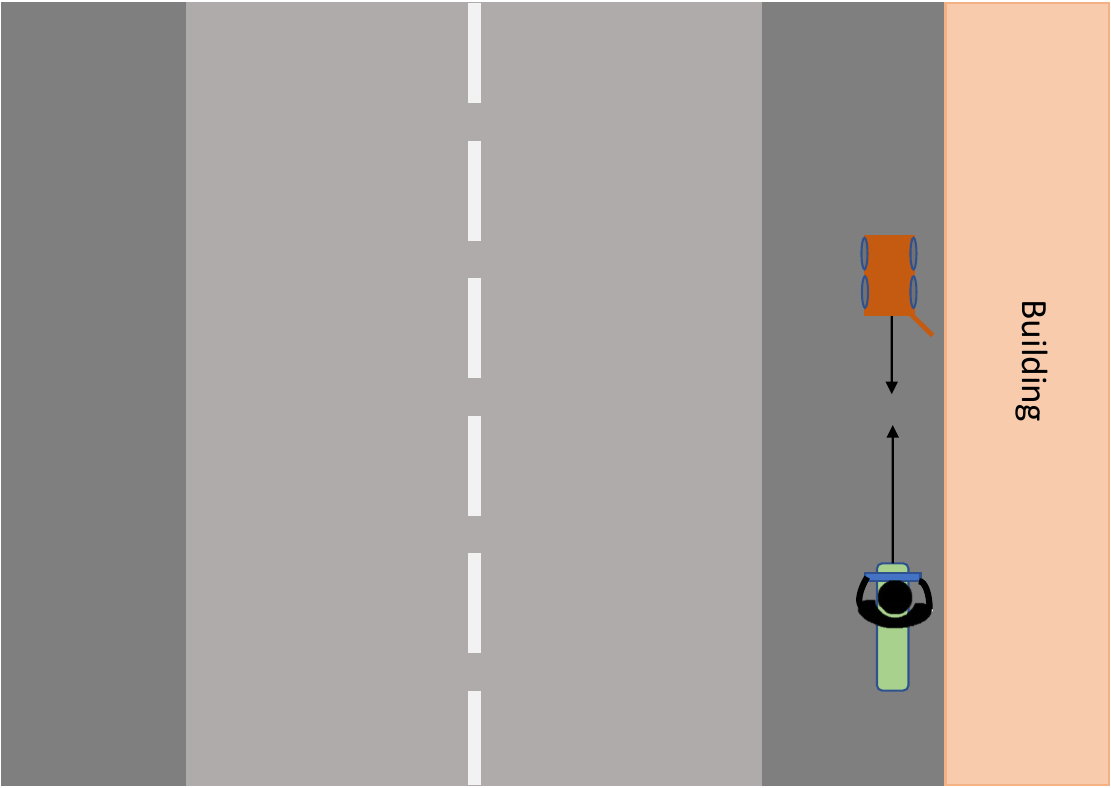}
        \caption{Scenario $S2$}\label{fig:s2}
    \end{subfigure}
    \hspace{20pt}
    \begin{subfigure}[b]{0.15\textwidth}
        \centering
        \includegraphics[width=\textwidth]{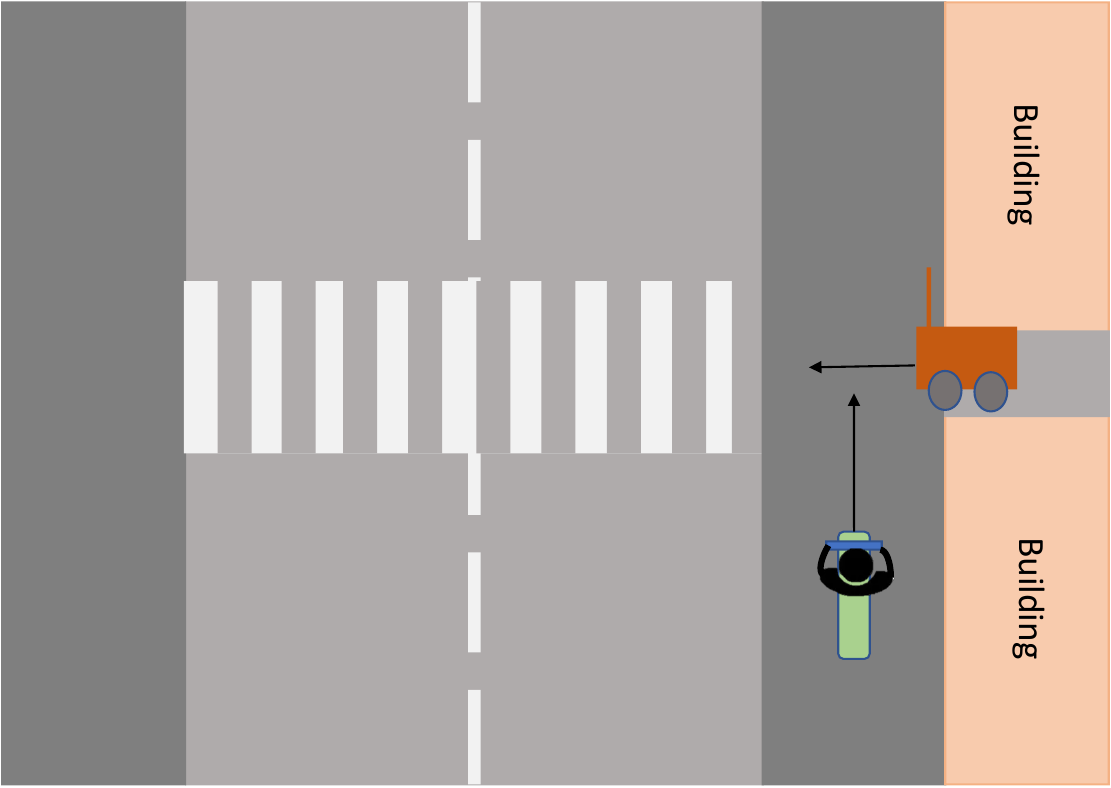}
        \caption{Scenario $S3$}\label{fig:s3}
    \end{subfigure}
    \hspace{20pt}
    \begin{subfigure}[b]{0.15\textwidth}
        \centering
        \includegraphics[width=\textwidth]{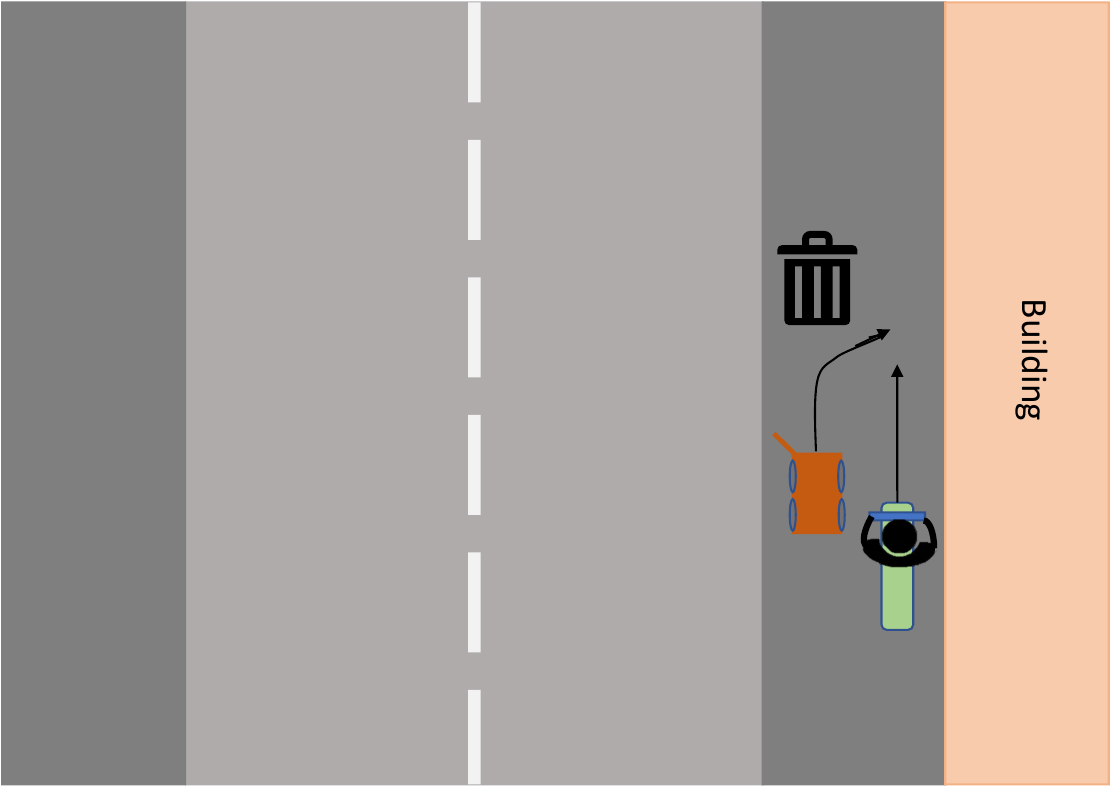}
        \caption{Scenario $S4$}\label{fig:s4}
    \end{subfigure}
    \caption{Four scenarios of ego-delivery robot interaction. In scenarios $S1$, $S2$, and $S3$, the roles of contributor and receiver are interchanged depending on who takes the yielding or unyielding action first. In scenario $S4$, the scooter acts as the contributor while the robot is the receiver. The robot and scooter switch locations in the robot-contributor scenario of $S4$.}
    \label{fig:scenario}  \vspace{-1.6em}
\end{figure*}
% \vspace{-5pt}
\subsection{Observational Study}
In our observational study, we focus on the interaction between a self-driving scooter as the AV and delivery robots as other road users. We investigate accommodative actions at the strategic level, consisting of yielding (as positive prosocial action) and unyielding actions during a conflict of path in an interaction. We specifically chose to examine sidewalk interactions, where there are no formal road rules regarding yielding or unyielding actions. This setting allows us to explore how these road users navigate shared spaces without predefined rules, relying instead on their accommodative actions when faced with potential path conflicts.

\textbf{Experiment Design:} We conducted a mixed design study, measuring the impact of delivery robot and scooter accommodative actions on user's well-being and trust. The study was designed as a $2 \text{(Other's action)}\times 2 \text{(Ego's action)}$ experiment, where the user rode a self-driving scooter and interacted with a delivery robot during the ride. Other's action had two levels: the robot yielding to the ego $O_+$ or unyielding to the ego $O_-$. Similarly, ego's action had two levels: the ego scooter yielding to the delivery robot $R_+$ or unyielding to the delivery robot $R_-$. This study was conducted using a Wizard of Oz experiment, where human operators controlled the scooter's movements to simulate strategic decision-making processes. This approach allowed us to focus on analyzing yielding and unyielding behaviors at the action level without the influence of specific trajectory-planning algorithms.

\textbf{Events and Scenarios:} Each participant interacts with two rides, with each ride comprising a sequence of two events: first event with delivery robot as the contributor (ego as the receiver) and second event with ego as the contributor (delivery robot as a receiver). This two-event sequence repeats in the second ride, with different scenarios. Four interaction scenarios were developed where the contributor's accommodative action is either yielding or unyielding. Figure~\ref{fig:scenario} displays the four scenarios ($S1$--$S4$) used in the study. The yielding actions in these scenarios only include actions such as stopping for the other to go first ($S1$ and $S3$), merging ($S4$), or changing the way for the other ($S2$). The order of scenarios across the four events was counterbalanced based on a Balanced Latin Square design. \looseness = -1

We have two independent variables with two levels each: robot's accommodative actions ($O_+$ and $O_{-}$) and ego's accommodative actions ($R_+$ and $R_-$). These variables are manipulated as both within-subjects and between-subjects factors. Given that each ride consists of an accommodative action by the robot followed by accommodative action by the scooter, there are 4 ($2\times2$) possible accommodative action combinations. To reduce the number of cases, the second ride has the same accommodative action from the robot as in the first ride; the scooter has all four permutations of accommodative actions across the two rides. This results in a total of $32$ combinations.
After each event, we measure the user's well-being and trust using a self-report questionnaire. During the second event in each ride, where the ego acts as a contributor, the user's intention toward the delivery robot is asked before the scooter exhibits its accommodative action. This allows to determine whether the user's intentions align with the scooter's action or not. \\
\textbf{Well-being and Trust Questionnaire:} To assess user's well-being, we used a modified version of the social well-being questionnaire \cite{radzyk2014validation} and made it situational rather than general and more applicable to our study scenario. The questionnaire was designed to measure well-being based on four factors: (1) positive relationship, (2) satisfaction with travel, (3) trust, and (4) general well-being. To calculate user's well-being, we average across all seven questions. Additionally, we asked a question specifically related to the user's trust in the self-driving scooter to measure trust independently (The details of the questionnaire used in this study are in the supplementary material.) To ensure a more consistent users' understanding of well-being and trust, we defined these concepts during the introduction of the user study to the participants. Moreover, users had access to the definitions of specific words (such as `trust') while answering the questions. 
% However, the study design allowed for individual interpretations of these concepts, acknowledging that capturing this diversity is vital for developing a robust and applicable model. \looseness=-1

\begin{figure}[t]
    \centering
    \includegraphics[width=.4\textwidth]{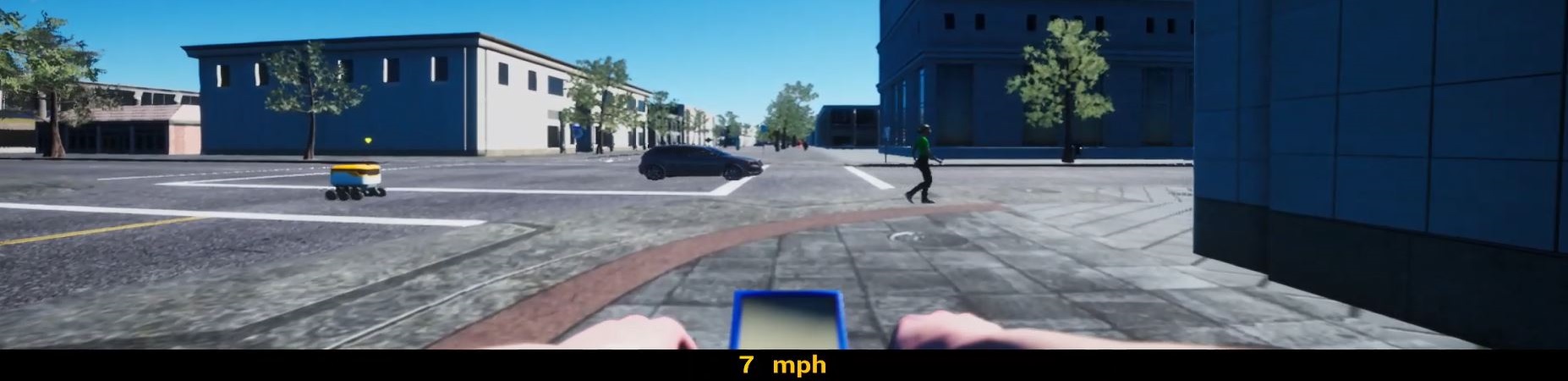}
    \caption{Web-based riding environment.}
    \label{fig:stimuli}\vspace{-1.8em}
\end{figure}
\textbf{User's intention: }To determine user's intention in the interaction where the ego is the contributor, we ask them a specific question: ``What action would you like your self-driving scooter to take regarding the delivery robot?'' Two options are given, with one implying yielding action and the other unyielding action. Based on the user's response, we assess whether the user's intention aligns with the action of the self-driving scooter in that particular event.

\textbf{Participants:} A total of $300$ participants were recruited via Prolific (\url{https://www.prolific.co/}), with the majority of participants being between $25$ and $55$ years old ($71.23\%$). Of the participants, $54\%$ identified as males, $44\%$ as females, and $2\%$ as others. All participants passed attention checks and indicated their commitment to thoughtfully answering survey questions. Participants were compensated $\$3.0$ for their approximately $25$-minute participation. All participants provided informed consent.

\textbf{Stimuli:}
The study was conducted online, and we used video recordings from a custom medium-fidelity driving simulator to simulate the scenarios. The simulated environments were created using Unreal Engine 4.27 (\url{https://www.unrealengine.com/}) with AirSim \cite{shah2018airsim}. To provide a realistic experience, the videos were recorded using one front-facing camera with $133$ degree horizontal field of view. Additionally, the scooter's speed information was overlaid at the bottom of the screen to enhance the experience (see Figure \ref{fig:stimuli}). The details of the study procedure, along with example recordings of events and scenarios are provided in the supplementary material. The supplementary materials, including the dataset and source code for the models discussed in the following sections, are available on GitHub at \url{https://github.com/honda-research-institute/wellbeing-trust-model}.
\subsection{Refinement of State Relationships}
Using the collected data, we examined statistical dependencies between latent states and observed variables to refine the structural relationships within the DBN models. Specifically, we applied statistical one-tail and two-tail t-tests to assess conditional independencies and validate the initial model structure.
Across all $32$ conditions of the study, we grouped the participants' data into different groups based on three factors: (1) robot's yielding vs. unyielding action, (2) scooter's yielding vs. unyielding action, and (3) alignment of participant’s intention with the scooter’s action. \\
\begin{figure}[t]
    \centering
    \includegraphics[width=.3\textwidth]{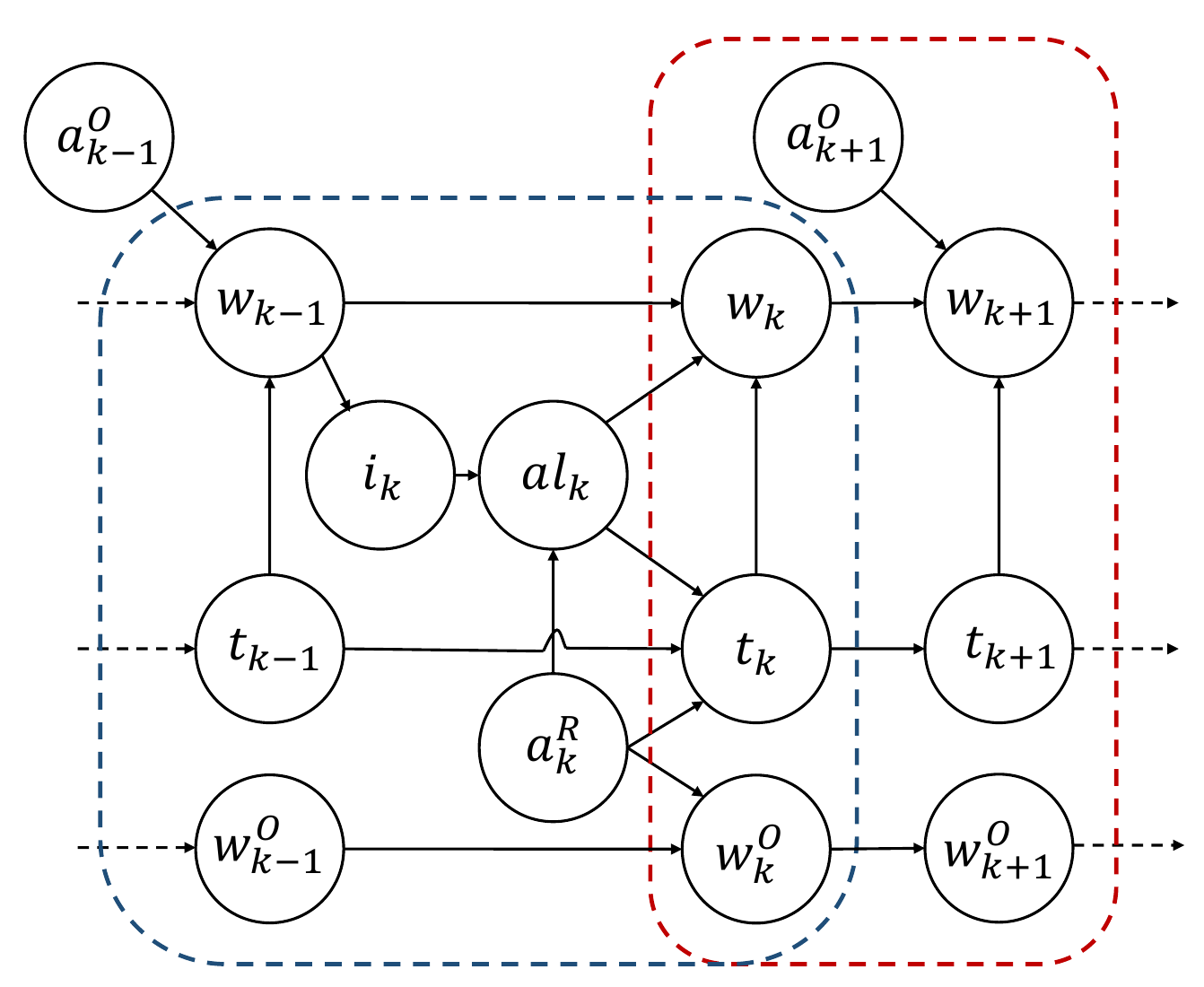}
    \caption{Final model structure of R-DBN (blue) and O-DBN (red)}
    \label{fig:dbn} \vspace{-1.8em}
\end{figure}
Data analysis was conducted to identify correlations between the scooter's action $a^R_k$, the user's intention alignment state $al_k$, robot delivery's action $a^O_k$, and the evolution of the user's latent states $x^E_k$. Among these, we identified statistically significant correlations with a p-value of $<0.05$. Specifically, we have the user's well-being is higher ($t(594.87) = 7.65$, $p < 0.0001^{(****)}$) when the delivery robot exhibits yielding action toward the ego compared to unyielding action. The user's trust is higher when the scooter exhibits yielding action than unyielding action toward others ($t(576.114) = 5.54$, $p < 0.0001^{(****)}$). The users with yielding intention toward others has higher well-being than those with unyielding intention ($t(317.49) = 2.12$, $p = 0.02$). The user's well-being and their trust in the scooter are higher when the scooter's action toward others is aligned with the user's intention compared to when it is not aligned ($t(577.26) = 7.64$, $p < 0.0001^{(****)}$). User's trust is positively correlated with well-being ($r(597) = 0.7058$,  $p < 0.0001^{(****)}$). Based on these findings, we adjusted the structure of the DBN models by setting the likelihood of relationships between variables without significant correlation to zero. To capture other's well-being, we assume that the effect of scooter's action on robot is the same as robot's actions on the user. To ensure that the model accurately reflects the observed dependencies in the data we adjusted the causal structure of the model. In particular, we performed a 5-fold cross-validation to evaluate candidate models and selected the best structure based on log-likelihood scores (SC-DBN: $-2854.90$ and RC-DBN: $-2698.20$ ). See Figure \ref{fig:dbn} for final model structure.

\subsection{Bayesian Parameter Estimation}
After finalizing the structure of the DBN models, we estimated the conditional probability distributions (CPDs) for each node using Bayesian parameter estimation. Given the dataset of observed state transitions, we computed the posterior distribution over CPDs using a Dirichlet prior.

To facilitate inference, we queried the propagated belief at event $k$ based on past experiences. To simplify CPDs, we scaled latent state values between $0$ and $1$ and discretized them into six bins (selected through ablation studies), allowing representation in a tabular format.

For discrete variables, the CPDs were represented as tabular distributions, where the conditional probability of a variable  $X$ given its parent variables $Pa(X)$ is given by \looseness = -1
% \vspace{-5pt}
\begin{eqnarray}
    P(X|Pa(X)) = \frac{N+\alpha}{\sum_X N+\alpha.k}
\end{eqnarray}

where $N$ is the observed count of transitions from the dataset, $\alpha$ is the Dirichlet prior, and $k$ is the number of possible states for each variable.

Using our dynamic Bayesian networks and available data, we estimated these CPDs with the Bayesian Parameter Estimator, applying a Dirichlet prior with a uniform $\alpha$. We implemented our DBN models using the pgmpy library in Python \cite{ankan2015pgmpy}.

By leveraging Bayesian parameter estimation and discretization, our DBN effectively captures probabilistic dependencies while enabling robust inference across different interaction scenarios.
\vspace{-5pt}
\subsection{Model Evaluation}
To evaluate the model’s performance in inferring key variables, we focused on assessing its accuracy in predicting user's \textit{well-being}, \textit{trust}, and \textit{intention}. Accuracy was measured as the proportion of correct inferences out of all the inferences made during the evaluation. We employed 5-fold cross-validation to assess the model’s generalization performance, repeating the cross-validation procedure for 100 iterations to ensure stable and reliable accuracy estimates.

The model achieved an accuracy of 77\% for inferring \textit{well-being}, 67\% for \textit{trust}, and 95\% for \textit{intention}, which demonstrates its effectiveness in capturing the dynamics of these important variables.

In addition to accuracy, we also evaluated how well the model could infer changes in these variables over time under different conditions. Specifically, we examined the evolution of \textit{well-being} $w_k$, \textit{trust} $t_k$, and \textit{other’s well-being} $w^O_k$ during 10 consecutive events of scooter interactions, where we fixed evidence and control variables such as \textit{action alignment} and the \textit{scooter’s action}. This approach allowed us to isolate the effects of action alignment and the scooter’s actions on the inferred variables.
\begin{figure*}[t]
    \centering
    \begin{subfigure}[b]{0.28\textwidth}
        \centering
        \includegraphics[width=\textwidth]{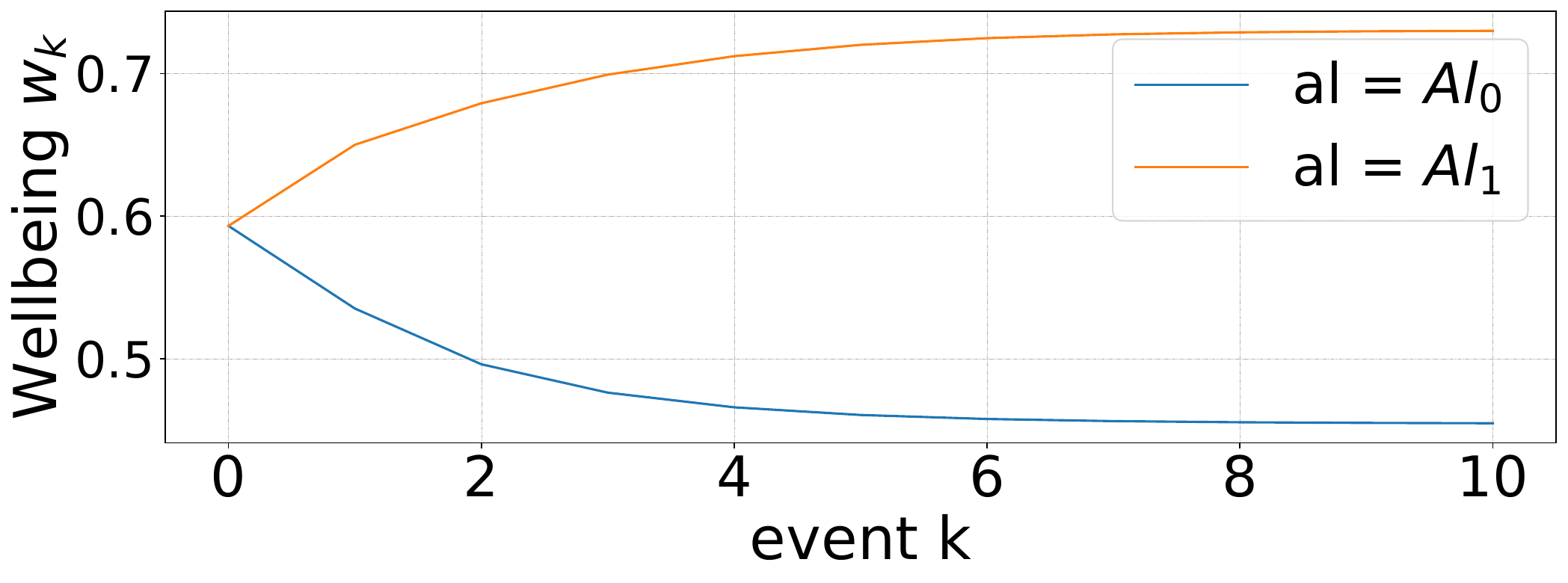}
        \caption{User's well-being.}\label{fig:inf1}
    \end{subfigure}
    \hfill
    \begin{subfigure}[b]{0.28\textwidth}
        \centering
        \includegraphics[width=\textwidth]{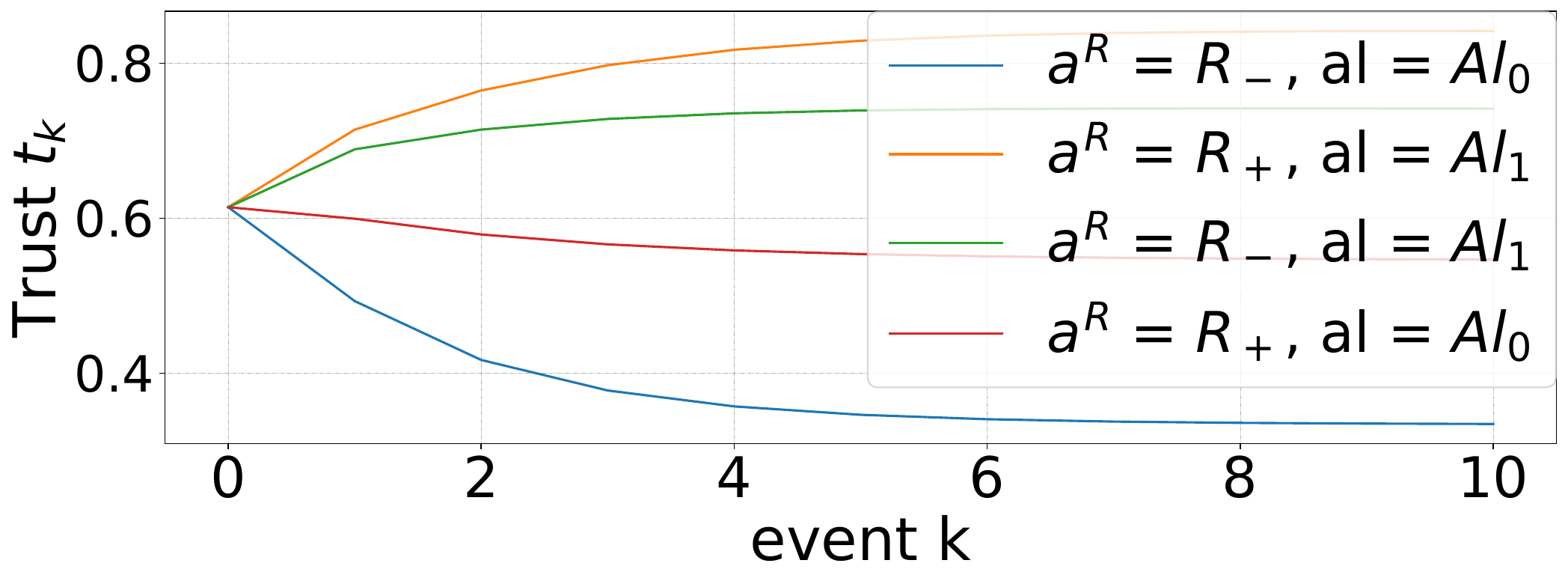}
        \caption{Trust.}\label{fig:inf2}
    \end{subfigure}
    \hfill
    \begin{subfigure}[b]{0.28\textwidth}
        \centering
        \includegraphics[width=\textwidth]{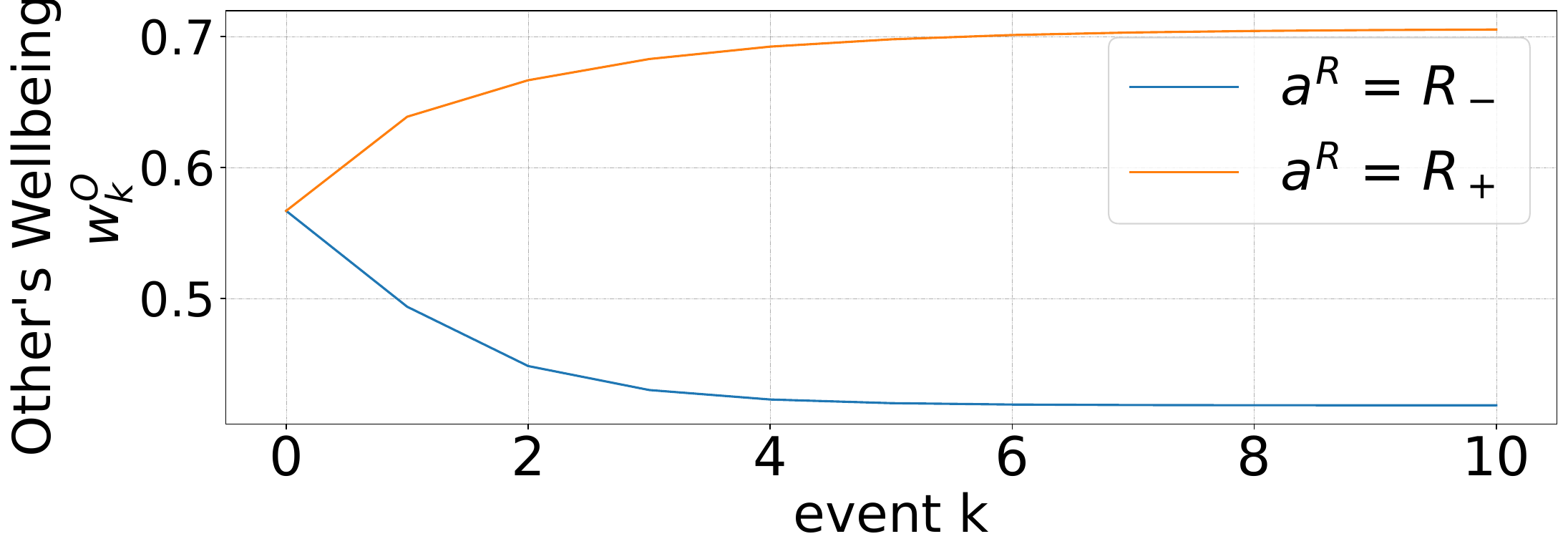}
        \caption{Other's well-being.}\label{fig:inf3}
    \end{subfigure}
    \caption{Expected values of inferred states over $10$ events using the model given the scooter's actions and action alignment.}
    \label{fig:infs}  
    \vspace{-1.8em}
\end{figure*}
To further assess the model’s performance, we analyzed how the inferred variables evolved across these 10 events. Figure \ref{fig:infs} illustrates the expected values for the \textit{user’s well-being} $w_k$, \textit{trust} $t_k$, and \textit{other’s well-being} $w_k^O$ over these events. 
As seen in the figure, the user's well-being increases over time when the scooter's action is aligned with the user's intention and decreases otherwise. Trust also shows dynamic changes depending on the action alignment and accommodative actions. Furthermore, other's well-being increases when the scooter takes yielding action and decreases when it takes unyielding action.

% As shown in the figure:

% \begin{itemize}
%     \item The \textit{user’s well-being} tends to increase over time when the scooter’s actions align with the user’s intentions, and it decreases when the actions do not align.
%     \item \textit{Trust} exhibits dynamic changes, reflecting the impact of action alignment and whether the scooter performs yielding or unyielding actions.
%     \item \textit{Other’s well-being} increases when the scooter adopts a yielding action and decreases when the scooter takes an unyielding action.
% \end{itemize}
These findings highlight the model’s capacity to infer and capture the dynamic relationships between the variables and the actions taken by the scooter, providing valuable insights into user behavior over time.

% We evaluated the model's accuracy in inferring users' well-being, trust, and intention. Accuracy was defined as the proportion of correctly inferred variables out of the total number of inferences made. We employed 5-fold cross-validation, and to ensure the accuracy estimates were stable and reliable, we repeated the cross-validation procedure for $100$ iterations. The model achieved accuracies of $77\%$ for well-being, $67\%$ for trust, and $95\%$ for intention, demonstrating its effectiveness in inferring these key variables.
% Our proposed model allows us to infer the changes of variables such as well-being, trust, and other's well-being over time. To evaluate the model's performance in variable inference, we considered 10 events of scooter contributor interactions while fixing the evidence and control variables such as action alignment and scooter's action. We then observed how the inferred variables changed given each condition. Figure \ref{fig:infs} presents the expected values of inferred user's well-being $w_k$, trust $t_k$, and other's well-being $w_k^o$ over $10$ events. As seen in the figure, the user's well-being increases over time when the scooter's action is aligned with the user's intention and decreases otherwise. Trust also shows dynamic changes depending on the action alignment and accommodative actions. Furthermore, other's well-being increases when the scooter takes yielding action and decreases when it takes unyielding action.

%%%%%%%%%%%%%%%%%%%%%%%%%%%%%%%%%%%%%%%%%%%%%%%%%%%%
\vspace{-6pt}
\section{Informed Decision Making}
% \begin{figure}[t]
%     \centering
%     \includegraphics[width=.3\textwidth]{images/CIM.jpg}
%     \caption{Causal Influence Diagram (CID) representation of the model for the first step, where the scooter aims to maximize user well-being}
%     \label{fig:CIM}
%     \vspace{-10pt}
% \end{figure}
One of the primary applications of the proposed model is to develop human-aware automation that can account for human cognitive states while making AV decisions. %Specifically, the proposed model can not only be used to predict trust, well-being, and intention but can also exploit the learned model structure for optimal decision-making. 
In this section, we will focus on the use of our model to determine the optimal policy or action for a self-driving scooter (AV) despite the uncertainty in the user's state variables. \looseness = -1

% \subsection{Casual Inference Model}
For optimal decisions, we need to define a utility function that we want to maximize given the model. We use causal inference modeling (CIM) to determine a policy that the self-driving scooter can adopt to optimize certain factors while taking uncertainty into account. CIM is a generalization of the Bayesian network that is used to represent decision-making processes under uncertainty. We transform our proposed dynamic Bayesian models into CIM with a casual inference diagram with chance nodes, utility nodes, and decision nodes. A utility node represents the outcome or value that a decision-maker is trying to optimize. A chance node represents an uncertain event or variable that can affect the outcome, while a decision node represents a point in the diagram where a decision-maker has control over the value of a variable. Depending on the factor that the scooter aims to optimize, the utility node may differ, such as user's well-being, user's trust, other's well-being, scooter's costs, or a trade-off between multiple factors. Additionally, when the generated policy pertains to the actions the scooter should take, the decision node represents the scooter's accommodative action, denoted as $a^R_k$. We used PyCID library in Python \cite{james_fox-proc-scipy-2021} to implement our casual inference modeling. We used the conditional probability distributions of each variable that we estimated using the data for our dynamic Bayesian models to build the CIM.
Given our proposed model and the utility function we want to optimize, we can use casual inference modeling to reason over the expected utility of each action and choose the action that maximizes the expected utility. Formally, given the evidence $ev$, the policy $\pi$ at event $k$ is given by
\begin{eqnarray}
\pi = \argmax_{a^R_k\in \{E_u, E_y\}} \{ E[U(a^R_k| ev)] \}   .
\end{eqnarray} 
Here, $E[U(a^R_k| ev)]$ represents the expected utility of taking the action $a^R_k$, which is computed as the sum over probability of all possible outcome states of $a^R_k$, i.e., $O_i(a^R_k)$, given the evidence and the action, and then multiplied by its corresponding utility function $U(O_i(a^R_k)|a^R_k)$ as
\begin{eqnarray}
E[U(a^R_k|& ev)] = \sum_i p(O_i(a^R_k)|ev , a^R_k)\times U(O_i(a^R_k)|a^R_k).
\end{eqnarray}
By defining an appropriate utility function, we can find the action that maximizes the expected utility. We analyze different policies that the scooter can adopt based on its objectives: (1) maximizing user's well-being ($U_k = w_k$), (2) maximizing user's trust ($U_k = t_k$), and (3) optimizing a trade-off of user's well-being, other's well-being, and the cost of its actions ($U_k = w_k + w_k^O + C(a^R_k)$). With these utilities, we can determine the optimal actions the scooter should take in different scenarios to achieve its desired objective.

\textbf{Maximizing User's Well-being:} We define the utility function as equal to the value of well-being. Using the model, we then determine the policy that maximizes user's well-being. In the absence of any evidence, the optimal policy for the scooter is to always take a yielding action. To further analyze the policy, we conducted an analysis of which nodes have a positive value of information (VOI) incentives. VOI incentives represent the expected increase in utility that can be achieved by acquiring additional information. VOI analysis shows that having evidence of previous well-being ($w_{k-1}$), trust ($t_{k-1}$), and current user intention ($i_k$) can lead to better reasoning of the optimal policy and consequently improve the utility. Thus, we analyze the optimal policy given the availability of evidence on these variables. Results show that if the scooter has access to information about the previous user's well-being, the optimal policy would be 
\begin{eqnarray}
a^R_k =       \begin{cases}
                R_- \quad \text{unyielding} &\quad\text{if } 0 \leq w_{k-1} \leq 0.18\\
                R_+ \quad\text{yielding} &\quad\text{\textit{otherwise}}    \hspace{5pt} .
            \end{cases}
\end{eqnarray} 
The optimal policy with information about the user's intention is to be aligned with user's intention, i.e.,
\begin{eqnarray}
a^R_k =       \begin{cases}
                R_- \quad \text{unyielding} &\quad\text{if } i_k = I_-\\
                R_+ \quad\text{yielding} &\quad\text{if } i_k = I_+    \hspace{5pt} .
            \end{cases}     
\end{eqnarray} 
Furthermore, the availability of previous trust information does not alter the policy when no evidence is available.\looseness=-1

\textbf{Maximizing User's Trust:}
We define the utility at event $k$ as the user's trust at that event. We identify that the previous user's well-being $w_{k-1}$, trust $t_{k-1}$, and current user's intention $i_k$ have positive VOI when the scooter aims to maximize user trust. When no evidence is available, the optimal policy is always to take the yielding action. However, if information about the user's intention is available, adopting the alignment policy can increase trust (similar to the case $U_k = w_k$), while having information about user's previous well-being and trust does not affect the optimal policy when no evidence is available.\looseness=-1

\begin{figure}[t]
    \centering
    \includegraphics[width=.25\textwidth]{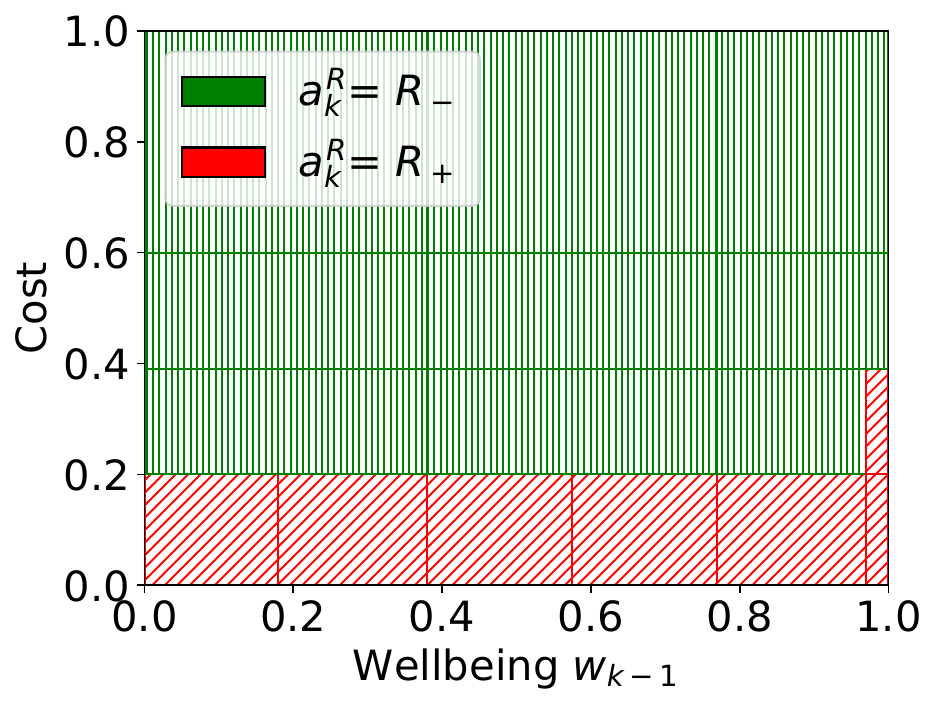}
    \vspace{-5pt}
    \caption{Optimal accommodative policy at event $k$. Sensitivity study results based on costs and information variables $w_{k-1}$.}
    \label{fig:policy} \vspace{-1.8em}
\end{figure}

\textbf{Optimizing the Trade-off between User's Well-being, Other's Well-being and Scooter's Cost:}
% \textbf{Balancing User Well-being, Others' Well-being, and Scooter Costs}
In this scenario, we consider that the cost of the scooter's action is different depending on whether it takes a yielding or an unyielding action. Since yielding actions usually involve changing direction or stopping, they are more costly (for example, more fuel spent) than unyielding actions. To simplify the problem, we assume that the cost difference is incorporated in the utility function, which becomes:
\begin{eqnarray}
 U_k =       \begin{cases}
                w_k + w^O_k &\quad\text{if } a^R_k = R_-\\
                w_k + w^O_k + C &\quad\text{if } a^R_k = R_+\\
            \end{cases}    
\end{eqnarray} 
The analysis shows that the cost of actions can significantly impact the utility and, thus, the policy. Therefore, we perform a sensitivity analysis of the cost values and examine how the policy changes according to different costs. We consider costs to be any value greater than or equal to zero. However, since the values of well-being range between $0$ and $1$, we find that the most critical costs are the ones that fall within this range. Figure \ref{fig:policy} illustrates how the policy changes with different information on the user's well-being. Analysis over other positive VOI evidence, including, trust, intention, and others' well-being, as well as in the absence of evidence shows that for costs less than $0.2$, the scooter should take yielding action and for more than that should take unyielding action.\looseness = -1

To sum up, we can see that if the cost difference between the two types of actions is large, the optimal action for the scooter is to always take unyielding actions. However, this would sacrifice the well-being of the user and others. Thus, when the cost difference is lower, we must balance the trade-off between well-being and cost to determine the optimal policy. For a cost difference of approximately $0.2$, the scooter's optimal policy is to mostly take yielding actions.% unless the user or others are in a low well-being state ($0 \leq w_{k-1} \leq 0.18$ and $0.18 < w^o_{k-1} \leq 0.38$)
However, for costs between $0.2$ and $0.4$, the scooter should mostly take unyielding actions unless the user or others are in a very high well-being state. Therefore, our proposed causal inference model allows us to make informed decisions about the best course of action for a scooter in order to achieve its objectives while balancing various factors.
\vspace{-7pt}
\section{Conclusions and Future Work}
Our paper proposes a computational model for autonomous vehicles that infers the cognitive states such as well-being, trust, and intentions of users and other road user's well-being to make informed decisions.
% a novel computational model for autonomous vehicles that can infer the cognitive states of both its user and other road users and uses this information to make informed decisions.
Our DBN models provide a structured representation of these states, enabling probabilistic inference. To learn the parameters of this model, we conducted an observational study to collect interaction data, which was then used to refine state relationships, estimate Bayesian parameters, and evaluate the model’s effectiveness. By integrating this model into AV decision-making, the system can account for well-being, trust, and cost trade-offs, leading to safer and more user-centered interactions with potential to inspire future research in autonomous vehicles.

Finally, while our proposed model provides a structured framework for inferring trust, well-being, and intention in human-AV interactions, the observational study conducted for parameter learning introduces certain limitations. Since the study consisted of only four interaction events per participant, the collected data may not fully capture the long-term evolution of trust and well-being. Future work will explore longitudinal data collection to better model these temporal dynamics.
Additionally, our study focused on dyadic interactions between a self-driving scooter (AV) and delivery robots in sidewalk environments without formal yielding rules. While this setting enabled controlled observation of accommodative actions, future work will expand the model to include more road users and mobility modes.
Despite these constraints, our approach marks progress toward human-aware automation. Using probabilistic inference, the model allows AVs to reason about trust and well-being, supporting safer, user-centered systems. Future research can enhance adaptability by incorporating domain-specific variables, contextual data, and broader interaction scenarios. 
\bibliographystyle{named}
\bibliography{ijcai24}

\begin{thebibliography}{}

\bibitem[\protect\citeauthoryear{Akash \bgroup \em et al.\egroup }{2020}]{akash2020toward}
Kumar Akash, Neera Jain, and Teruhisa Misu.
\newblock Toward adaptive trust calibration for level 2 driving automation.
\newblock In {\em ICMI}, pages 538--547, 2020.

\bibitem[\protect\citeauthoryear{Ankan and Panda}{2015}]{ankan2015pgmpy}
Ankur Ankan and Abinash Panda.
\newblock pgmpy: Probabilistic graphical models using python.
\newblock In {\em SCIPY}. Citeseer, 2015.

\bibitem[\protect\citeauthoryear{Azevedo-Sa \bgroup \em et al.\egroup }{2020}]{azevedo2020context}
Hebert Azevedo-Sa, Suresh~Kumaar Jayaraman, X~Jessie Yang, Lionel~P Robert, and Dawn~M Tilbury.
\newblock Context-adaptive management of drivers’ trust in automated vehicles.
\newblock {\em RA-L}, 5(4):6908--6915, 2020.

\bibitem[\protect\citeauthoryear{Bowen and Smith}{2019}]{bowen2019drive}
Louise Bowen and Andrew Smith.
\newblock Drive better, feel better: Examining associations between well-being and driving behaviour in students.
\newblock {\em Advances in Social Sciences Research Journal}, 6(2), 2019.

\bibitem[\protect\citeauthoryear{Deo and Trivedi}{2019}]{deo2019looking}
Nachiket Deo and Mohan~M Trivedi.
\newblock Looking at the driver/rider in autonomous vehicles to predict take-over readiness.
\newblock {\em IEEE T-IV}, 5(1):41--52, 2019.

\bibitem[\protect\citeauthoryear{Ettema \bgroup \em et al.\egroup }{2011}]{ettema2011satisfaction}
Dick Ettema, Tommy G{\"a}rling, Lars Eriksson, Margareta Friman, Lars~E Olsson, and Satoshi Fujii.
\newblock Satisfaction with travel and subjective well-being: Development and test of a measurement tool.
\newblock {\em Trans. Research Part F: Traffic Psychology and Behaviour}, 14(3):167--175, 2011.

\bibitem[\protect\citeauthoryear{Friman \bgroup \em et al.\egroup }{2013}]{friman2013psychometric}
Margareta Friman, Satoshi Fujii, Dick Ettema, Tommy G{\"a}rling, and Lars~E Olsson.
\newblock Psychometric analysis of the satisfaction with travel scale.
\newblock {\em Transportation Research Part A: Policy and Practice}, 48:132--145, 2013.

\bibitem[\protect\citeauthoryear{Guhe \bgroup \em et al.\egroup }{2005}]{guhe2005non}
Markus Guhe, Wayne~D Gray, Michael~J Schoelles, Wenhui Liao, Zhiwei Zhu, and Qiang Ji.
\newblock Non-intrusive measurement of workload in real-time.
\newblock In {\em HFES}, volume~49, pages 1157--1161. SAGE Publications Sage CA: Los Angeles, CA, 2005.

\bibitem[\protect\citeauthoryear{Guo \bgroup \em et al.\egroup }{2020}]{guo2020modeling}
Yaohui Guo, Chongjie Zhang, and X~Jessie Yang.
\newblock Modeling trust dynamics in human-robot teaming: A bayesian inference approach.
\newblock In {\em CHI-Extended Abstracts}, pages 1--7, 2020.

\bibitem[\protect\citeauthoryear{Halkola \bgroup \em et al.\egroup }{2019}]{halkola2019towards}
Eija Halkola, Lauri Lov{\'e}n, Marta Cortes, Ekaterina Gilman, and Susanna Pirttikangas.
\newblock Towards measuring well-being in smart environments.
\newblock In {\em UbiComp/ISWC}, pages 1166--1169, 2019.

\bibitem[\protect\citeauthoryear{Harris \bgroup \em et al.\egroup }{2014}]{harris2014prosocial}
Paul~B Harris, John~M Houston, Jose~A Vazquez, Janan~A Smither, Amanda Harms, Jeffrey~A Dahlke, and Daniel~A Sachau.
\newblock The prosocial and aggressive driving inventory (padi): A self-report measure of safe and unsafe driving behaviors.
\newblock {\em Accident Analysis \& Prevention}, 72:1--8, 2014.

\bibitem[\protect\citeauthoryear{Hu \bgroup \em et al.\egroup }{2013}]{hu2013negative}
Tian-Yi Hu, Xiaofei Xie, and Jie Li.
\newblock Negative or positive? the effect of emotion and mood on risky driving.
\newblock {\em TRANSPORT RES F-TRAF}, 16:29--40, 2013.

\bibitem[\protect\citeauthoryear{Isler and Newland}{2017}]{isler2017life}
Robert~B Isler and Samantha~A Newland.
\newblock Life satisfaction, well-being and safe driving behaviour in undergraduate psychology students.
\newblock {\em TRANSPORT RES F-TRAF}, 47:143--154, 2017.

\bibitem[\protect\citeauthoryear{{J}ames {F}ox \bgroup \em et al.\egroup }{2021}]{james_fox-proc-scipy-2021}
{J}ames {F}ox, {T}om {E}veritt, {R}yan {C}arey, {E}ric {L}anglois, {A}lessandro {A}bate, and {M}ichael {W}ooldridge.
\newblock {P}y{C}{I}{D}: {A} {P}ython {L}ibrary for {C}ausal {I}nfluence {D}iagrams.
\newblock In {M}eghann {A}garwal, {C}hris {C}alloway, {D}illon {N}iederhut, and {D}avid {S}hupe, editors, {\em SCIPY}, pages 43 -- 51, 2021.

\bibitem[\protect\citeauthoryear{Koppol \bgroup \em et al.\egroup }{2021}]{koppol2021interaction}
Pallavi Koppol, Henny Admoni, and Reid~G Simmons.
\newblock Interaction considerations in learning from humans.
\newblock In {\em IJCAI}, pages 283--291, 2021.

\bibitem[\protect\citeauthoryear{Lee and See}{2004}]{lee2004trust}
John~D. Lee and Katrina~A. See.
\newblock Trust in automation: {{Designing}} for appropriate reliance.
\newblock {\em Human Factors}, 46(1):50--80, 2004.

\bibitem[\protect\citeauthoryear{Liang and Lee}{2014}]{liang2014hybrid}
Yulan Liang and John~D Lee.
\newblock A hybrid bayesian network approach to detect driver cognitive distraction.
\newblock {\em Transp. Res. Part C Emerg. Technol.}, 38:146--155, 2014.

\bibitem[\protect\citeauthoryear{Liang \bgroup \em et al.\egroup }{2007}]{liang2007nonintrusive}
Yulan Liang, John~D Lee, and Michelle~L Reyes.
\newblock Nonintrusive detection of driver cognitive distraction in real time using bayesian networks.
\newblock {\em Transportation research record}, 2018(1):1--8, 2007.

\bibitem[\protect\citeauthoryear{Luo \bgroup \em et al.\egroup }{2019}]{luo2019toward}
Ruikun Luo, Yifan Wang, Yifan Weng, Victor Paul, Mark~J Brudnak, Paramsothy Jayakumar, Matt Reed, Jeffrey~L Stein, Tulga Ersal, and X~Jessie Yang.
\newblock Toward real-time assessment of workload: A bayesian inference approach.
\newblock In {\em HFES}, volume~63, pages 196--200. SAGE Publications Sage CA: Los Angeles, CA, 2019.

\bibitem[\protect\citeauthoryear{Luo \bgroup \em et al.\egroup }{2021}]{luo2021workload}
Ruikun Luo, Yifan Weng, Yifan Wang, Paramsothy Jayakumar, Mark~J Brudnak, Victor Paul, Vishnu~R Desaraju, Jeffrey~L Stein, Tulga Ersal, and X~Jessie Yang.
\newblock A workload adaptive haptic shared control scheme for semi-autonomous driving.
\newblock {\em Accident Analysis \& Prevention}, 152:105968, 2021.

\bibitem[\protect\citeauthoryear{Mahmood \bgroup \em et al.\egroup }{2024}]{mahmood2024designing}
Syed Hasan~Amin Mahmood, Zhuoran Lu, and Ming Yin.
\newblock Designing behavior-aware ai to improve the human-ai team performance in ai-assisted decision making.
\newblock In {\em Proceedings of the Thirty-Third International Joint Conference on Artificial Intelligence}, pages 3106--3114, 2024.

\bibitem[\protect\citeauthoryear{Mehrotra \bgroup \em et al.\egroup }{2023}]{mehrotra2023wellbeing}
Shashank Mehrotra, Zahra Zahedi, Teruhisa Misu, and Kumar Akash.
\newblock {Wellbeing in Future Mobility: Toward AV Policy Design to Increase Wellbeing through Interactions}.
\newblock In {\em ITSC}, 2023.

\bibitem[\protect\citeauthoryear{Nouvian}{2023}]{nouvian2023paris}
Tom Nouvian.
\newblock In paris referendum, 89\% of voters back a ban on electric scooters, Apr 2023.

\bibitem[\protect\citeauthoryear{Ong \bgroup \em et al.\egroup }{2019}]{ong2019computational}
Desmond~C Ong, Jamil Zaki, and Noah~D Goodman.
\newblock Computational models of emotion inference in theory of mind: A review and roadmap.
\newblock {\em TOPICS}, 11(2):338--357, 2019.

\bibitem[\protect\citeauthoryear{Radzyk}{2014}]{radzyk2014validation}
JJ~Radzyk.
\newblock Validation of a new social well-being questionnaire.
\newblock {B.S.} thesis, University of Twente, 2014.

\bibitem[\protect\citeauthoryear{Sauer \bgroup \em et al.\egroup }{2019}]{sauer2019empirical}
Vanessa Sauer, Alexander Mertens, Verena Nitsch, and Jens~Dietmar Reuschel.
\newblock An empirical investigation of measures for well-being in highly automated vehicles.
\newblock In {\em AutomotiveUI: Adjunct Proceedings}, pages 369--374, 2019.

\bibitem[\protect\citeauthoryear{Seligman and others}{2002}]{seligman2002positive}
Martin~EP Seligman et~al.
\newblock Positive psychology, positive prevention, and positive therapy.
\newblock {\em Handbook of positive psychology}, 2(2002):3--12, 2002.

\bibitem[\protect\citeauthoryear{Shah \bgroup \em et al.\egroup }{2018}]{shah2018airsim}
Shital Shah, Debadeepta Dey, Chris Lovett, and Ashish Kapoor.
\newblock Airsim: High-fidelity visual and physical simulation for autonomous vehicles.
\newblock In {\em Field and Service Robotics: Results of the 11th International Conference}, pages 621--635. Springer, 2018.

\bibitem[\protect\citeauthoryear{Soh \bgroup \em et al.\egroup }{2020}]{soh2020multi}
Harold Soh, Yaqi Xie, Min Chen, and David Hsu.
\newblock Multi-task trust transfer for human--robot interaction.
\newblock {\em IJRR}, 39(2-3):233--249, 2020.

\bibitem[\protect\citeauthoryear{Stocker and Shaheen}{2017}]{stocker2017shared}
Adam Stocker and Susan Shaheen.
\newblock Shared automated vehicles: Review of business models.
\newblock International Transport Forum Discussion Paper 2017-09, Paris, 2017.

\bibitem[\protect\citeauthoryear{Wu \bgroup \em et al.\egroup }{2022}]{wu2022toward}
Tong Wu, Enna Sachdeva, Kumar Akash, Xingwei Wu, Teruhisa Misu, and Jorge Ortiz.
\newblock Toward an adaptive situational awareness support system for urban driving.
\newblock In {\em IV}, pages 1073--1080. IEEE, 2022.

\bibitem[\protect\citeauthoryear{Xu and Dudek}{2015}]{xu2015optimo}
Anqi Xu and Gregory Dudek.
\newblock Optimo: Online probabilistic trust inference model for asymmetric human-robot collaborations.
\newblock In {\em HRI}, pages 221--228. IEEE, 2015.

\bibitem[\protect\citeauthoryear{Xu and Dudek}{2016}]{xu2016maintaining}
Anqi Xu and Gregory Dudek.
\newblock Maintaining efficient collaboration with trust-seeking robots.
\newblock In {\em IROS}, pages 3312--3319. IEEE, 2016.

\bibitem[\protect\citeauthoryear{Zahedi \bgroup \em et al.\egroup }{2022}]{zahedi2022modeling}
Zahra Zahedi, Sarath Sreedharan, Mudit Verma, and Subbarao Kambhampati.
\newblock Modeling the interplay between human trust and monitoring.
\newblock In {\em HRI}, pages 1119--1123. IEEE, 2022.

\bibitem[\protect\citeauthoryear{Zahedi \bgroup \em et al.\egroup }{2023}]{zahedi2023trust}
Zahra Zahedi, Mudit Verma, Sarath Sreedharan, and Subbarao Kambhampati.
\newblock Trust-aware planning: Modeling trust evolution in iterated human-robot interaction.
\newblock In {\em HRI}, pages 281--289, 2023.

\bibitem[\protect\citeauthoryear{Zhuge and Zhuge}{2020}]{zhuge2020toward}
Hai Zhuge and Hai Zhuge.
\newblock Toward cyber-physical-social science.
\newblock {\em Cyber-Physical-Social Intelligence: On Human-Machine-Nature Symbiosis}, pages 341--351, 2020.

\end{thebibliography}

\newpage
\onecolumn 

\section*{Appendix A -- Wellbeing and Trust Questionnaire}
To assess user's well-being, we used a modified version of the social well-being questionnaire (Q1--Q7 in Table~\ref{tab:questionnaire}) \cite{radzyk2014validation} to make it situational rather than general and more applicable to our study scenario. Our questionnaire was designed to measure well-being based on four factors: (1) positive relationship, (2) satisfaction with travel \cite{friman2013psychometric}, (3) trust, and (4) general well-being (see Table \ref{tab:questionnaire}). \looseness = -1

For positive relationship, we included two questions to assess the user's perception of their relationship with others, one asking about their relationship toward others (Q1) and the other is asking about others' relationship toward them (Q2). To measure satisfaction with travel, we selected one question from each of the three factors in the Satisfaction with Travel Scale \cite{friman2013psychometric}: positive activation (Q3), positive deactivation (Q4), and cognitive evaluation (Q5). To assess trust as a factor of well-being, we included a question about the user's trust in others  (Q7). Additionally, we asked a question specifically related to the user's trust in the self-driving scooter to measure trust independently (Q8). Finally, we included a question to measure the user's overall sense of well-being after each interaction (Q6).
To calculate user's well-being, we average over the response values for Q1--Q7, and for user's trust in scooter, we use the response value for Q8. To ensure a more consistent users' understanding of well-being and trust, we defined these concepts during the introduction of the user study to the participants. Moreover, users had access to the definition of different words during the study while answering the questions. However, the study design allowed for individual interpretations of these concepts, acknowledging that capturing this diversity is vital for developing a robust and applicable model.
\begin{table}[ht]
    \centering
     \caption{Well-being (1--7) and Trust (8) questionnaire}
    \resizebox{0.7\columnwidth}{!}{%
    \begin{tabular}{p{13cm}l}
    \toprule
    \textbf{1. Positive relationship (me to others):} Based on the current interaction, I am content with the relation with other robot.\\
    \textbf{2.	Positive relationship (others to me):} Based on the current interaction, I think the delivery robot around me handle others in a positive manner.\\
    \textbf{3.	Satisfaction (positive activation):} During my current travel event, I was worried/confident.\\
    \textbf{4.	Satisfaction (positive deactivation):} During my current travel event, I was tired/alert.\\
    \textbf{5.	Satisfaction (cognitive evaluation):} My current travel event worked poorly/worked well.\\
    \textbf{6.	Well-being:} This travel event contributes to my well-being.\\
    \textbf{7.	Trust (in others):} Based on the current interaction, I trust robots in my surrounding.\\
    \textbf{8.	Trust (in scooter):} Based on the current interaction, I trust my self-driving scooter.\\
    \bottomrule 
    \end{tabular}}
    \label{tab:questionnaire}
\end{table}

\newpage
\section*{Appendix B -- Observational Study Procedure}

\begin{figure*}[ht]
    \centering
    \includegraphics[width=\columnwidth]{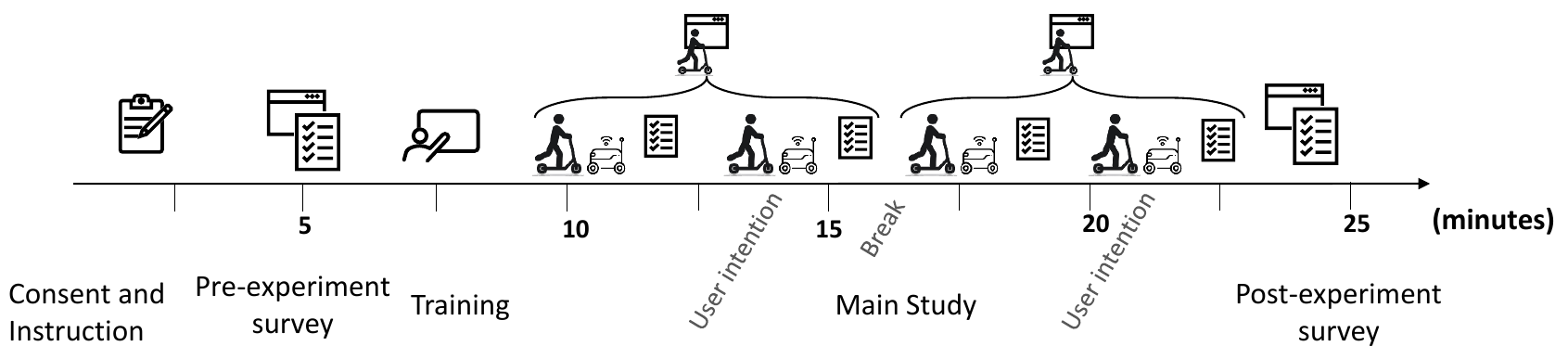}
    \caption{Study procedure overview}
    \label{fig:procedure}

\end{figure*}

As shown in Figure \ref{fig:procedure}, upon providing consent, each participant is directed to the instructions and the pre-experiment surveys: 
% the Warwick-Edinburgh Mental Well-being Scale (WEMWBS) \cite{tennant2007warwick} and 
a positive relationships and well-being questionnaire in a general format. Participants then undergo a 3-minute training session on the web-based driving simulation, which includes an introduction to speed, other sidewalk and road users, how they might interact with delivery robots, sample surveys (which are different from the main study), and a question on intention. Participants also learn how the scooter communicates the situation in a descriptive voice to the user, as well as how to indicate their intent to brake or decelerate using the space bar. Participants are required to complete two rides, each consisting of two-event interaction scenarios. After each interaction event, participants answer questions related to their well-being and trust. The participants expressed their level of agreement or disagreement with the questions using a 7-point Likert scale, ranging from ``strongly disagree'' to ``strongly agree''. In the second interaction event of each ride, before they see the scooter's accommodative action toward the robot, they are asked to indicate their preferred action for their self-driving scooter. Note that the actual scooter action is independent of the participant's response. After completing both rides, participants answer questions about their demographic information and previous experience with autonomous features and vehicles. They are then compensated for their participation. 

\end{document}